\documentclass[a4paper, 10pt, conference]{ieeeconf}      

\usepackage{FG2024}
\usepackage{graphicx}
\usepackage{amsmath}
\usepackage{arydshln} 
\usepackage{url}
\usepackage{multirow}
\usepackage{color}
\usepackage{adjustbox}
\usepackage[font=small,skip=2pt]{caption}
\newcommand{\cmmnt}[1]{}
\newcommand{\RE}{\mathrm{Re}}
\newcommand{\IM}{\mathrm{Im}}
\usepackage{amssymb}
\usepackage{pifont}

\FGfinalcopy 

\IEEEoverridecommandlockouts                              
\def\FGPaperID{157} 

\title{\LARGE \bf
Deep adaptative spectral zoom for improved remote heart rate estimation
}

\author{\parbox{16cm}{\centering
{\large Joaquim Comas$^1$, Adrià Ruiz$^2$, Federico Sukno$^1$}\\
{\normalsize $^1$ Department of Information and Communication Technologies, Pompeu Fabra University, Barcelona, Spain \\ $^2$ Seedtag, Madrid, Spain}}} 

\begin{document}

\maketitle
\begin{abstract}
Recent advances in remote heart rate measurement,  motivated by data-driven approaches, have notably enhanced accuracy. However, these improvements primarily focus on recovering the rPPG signal, overlooking the implicit challenges of estimating the heart rate (HR) from the derived signal. While many methods employ the Fast Fourier Transform (FFT) for HR estimation, the performance of the FFT is inherently affected by a limited frequency resolution. In contrast, the Chirp-Z Transform (CZT), a generalization form of FFT, can refine the spectrum to the narrow-band range of interest for heart rate, \cmmnt{(typically between 0.66 and 4 Hz)} providing improved frequential resolution and, consequently, more accurate estimation. This paper presents the advantages of employing the CZT for remote HR estimation and introduces a novel data-driven adaptive CZT estimator. The objective of our proposed model is to tailor the CZT to match the characteristics of each specific dataset sensor, facilitating a more optimal and accurate estimation of HR from the rPPG signal without compromising generalization across diverse datasets. This is achieved through a Sparse Matrix Optimization (SMO). We validate the effectiveness of our model through exhaustive evaluations on three publicly available datasets —UCLA-rPPG, PURE, and UBFC-rPPG— employing both intra- and cross-database performance metrics. The results reveal outstanding heart rate estimation capabilities, establishing the proposed approach as a robust and versatile estimator for any rPPG method.
\end{abstract}

\section{Introduction}
\label{sec:intro}

Recently, the research community has increasingly focused on the camera-based measurement of human physiological signals and their potential applications  \cite{ronca2021video, benezeth2018remote, liu20163d}, particularly in the extraction and analysis of vital signs such as the heart rate (HR), heart rate variability (HRV), respiration rate (RR), oxygen saturation (SpO2), and blood volume pulse (BVP). Among these vital signs, HR has been the most extensively studied 
due to its relevance to health and human-computer interaction applications, e.g. by using HR to complement facial expressions for the analysis of human emotions \cite{jung2019utilizing,comas2020end}\cmmnt{due to its intrinsic relation with the photoplethysmography (PPG) signal and its versatile applications}. Indeed, the majority of metrics used to evaluate the performance of remote photoplethysmography (rPPG) methods are HR-based, including mean absolute error (MAE), root mean square error (RMSE), and Pearson's correlation (R), all derived from the rPPG signal. 

\begin{figure}
    \centering
\includegraphics[width=85mm,height=70mm]{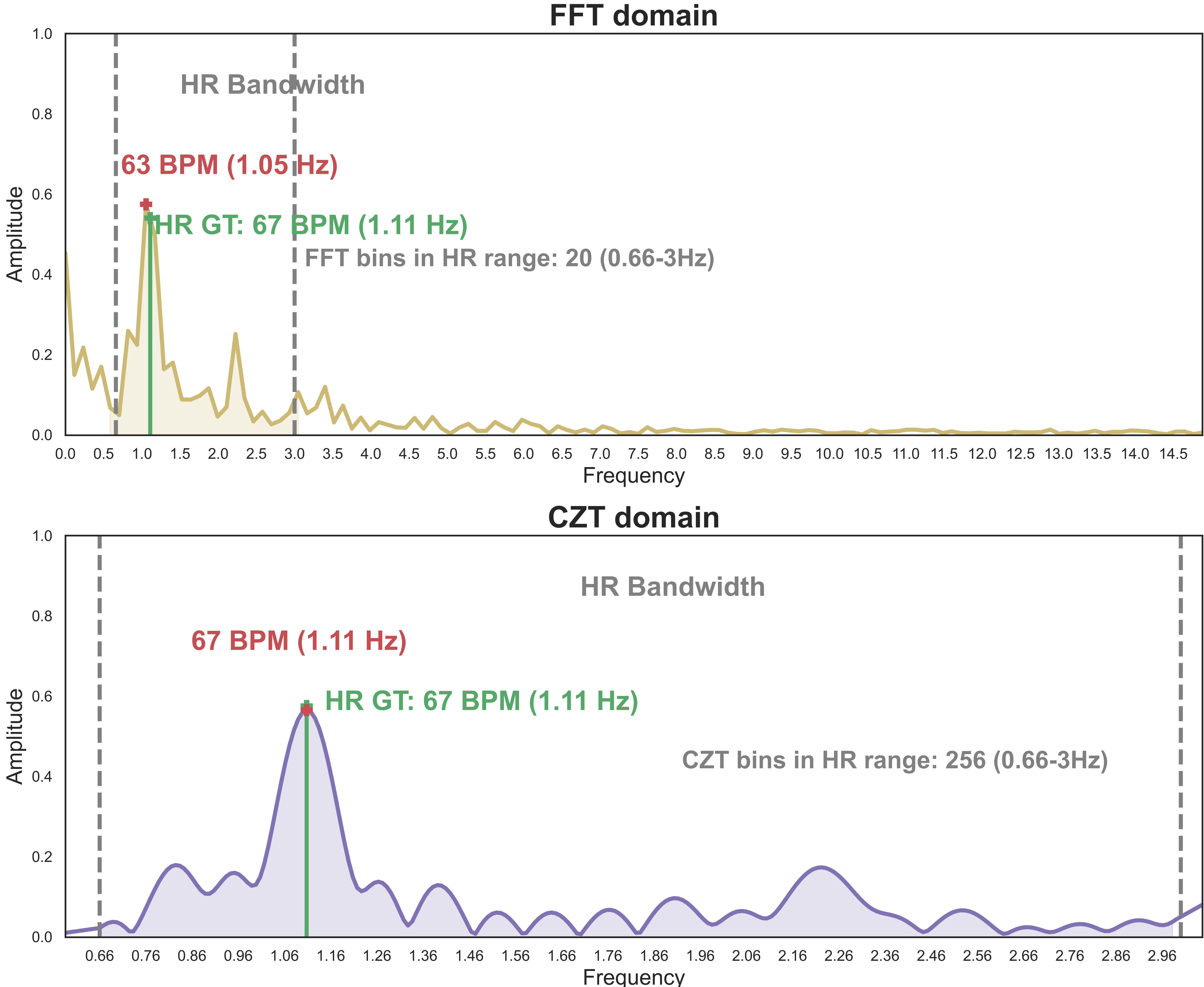}
    \caption{Frequency spectrum comparison between FFT  (above) and CZT (below) for a UCLA-rPPG PPG signal sample (8.5 sec windows). In red is denoted the predicted HR while green represents the HR ground truth.}
    \label{fig:FFTvsCZT}
\end{figure}

Conventional rPPG methods involve a two-stage process: initial signal extraction based on rPPG principles and subsequent HR calculation using signal processing. While recent progress has been made in the first stage \cite{chen2018deepphys,niu2019rhythmnet,yu2023physformer++}, the second stage relies heavily on traditional hand-crafted methods and remains largely unexplored. Three main approaches to derive HR from rPPG are distinguished: in the temporal domain, peak detection algorithms like Pan-Tompkins \cite{pan1985real} are commonly used, providing acceptable HR estimates, especially for ECG. However, rPPG signals can often be very noisy, requiring post-processing and manual fine-tuning, making this method challenging. In the frequency domain, the FFT is widely used for HR estimation. Despite its precision, its range is fixed by the Nyquist frequency, which is well above the interest range for HR, under a uniform spectrum sampling. Thus, the frequency resolution of the FFT is also fixed and often sub-optimal for HR estimation, e.g. very few frequency bins fall within the range of interest. 
Recent data-driven approaches propose end-to-end models, but they face challenges like convergence issues due to realistic rPPG-irrelevant noise \cite{yu2023physformer++} or overlook the ordinal nature of HR outputs, as seen in other deep learning problems like facial age estimation \cite{cao2020rank}.

In this paper, we propose the use of the Chirp-Z Transform (CZT) \cite{rabiner1969chirp} for the task of remote HR estimation. Originating in 1969, this method serves as a generalization of the widely used Discrete Fourier Transform (DFT). With its unique capabilities, it allows for a flexible choice of the spectral resolution\cmmnt{high-resolution narrow-band analysis}, rendering it particularly well-suited for HR estimation—a superiority that we will demonstrate in section \ref{sec:experiments} over traditional FFT estimation. 
Indeed, the CZT can change both the resolution and the range of frequencies to be covered, thus allowing to focus on a particular range of interest. This is illustrated in Figure 1. This feature also enables more precise estimations using short temporal windows of the rPPG signal, making it particularly suitable for continuous heart rate (HR) evaluation, as considered in the recent Vision-for-Vitals challenge (V4V) \cite{revanur2021first}. Most existing video-based physiological recovery methods have traditionally focused on predicting HR values over large window intervals (e.g., 30 seconds or video-level). While this evaluation protocol may provide unbiased performance for these methods, it limits their effectiveness in real-time applications, such as computing Heart Rate Variability (HRV) or diagnosing Atrial fibrillation \cite{pereira2020photoplethysmography}, where the CZT could offer a viable solution for these real-time HR applications.

Finally, inspired by previous works \cite{velik2008discrete, lee2021fnet} formulating the FFT computation as a neural network, we introduce a novel deep CZT estimator. In particular, we formulate the CZT as a set of fully differentiable network layers, allowing to fine-tune its parameters and thus, capturing better the relationship between the rPPG signal and the HR measurement sensor of the considered dataset. 
Although the goal is to improve the estimation by learning the characteristics of each HR sensor, we also introduce a Sparse Matrix Optimization (SMO) regularization loss to retain the generalization ability and structure of the standard CZT, balancing the deviations required to adapt to the considered sensor without over-fitting.
Interestingly, we find that the performance improvement of the proposed CZT is reproducible for different rPPG signal extraction methods, as we will empirically show in section \ref{subsec:sota}. This allows it to be integrated within any type of rPPG method, be it handcrafted or data-driven, and accommodates different temporal window sizes, enabling continuous HR estimation using the frequency domain. 


\subsection{Contribution}


In this work, we showcase the advantages of incorporating the Chirp-Z Transform for remote HR estimation. Our contribution extends to the introduction of a novel deep-learning CZT estimator module, specially designed to enhance HR estimation in rPPG conditions. The core objective of our proposed HR estimator is to acquire a customized variation of the classical CZT, adapting to the distinctive characteristics of each HR dataset sensor. To train our model effectively, we also introduce a novel Sparse Matrix Optimization (SMO) loss to not only facilitate robust generalization within its original dataset but also promote adaptability across diverse datasets by constraining the adaptation of the data-driven CZT. Another notable feature of the presented CZT estimator module is its flexibility to be integrated into any rPPG approach, providing a more precise HR estimator regardless of the adopted rPPG method. To validate our approach, comprehensive experiments are conducted using 3 publicly available datasets: UCLA-rPPG. PURE and UBFC-rPPG, and including both intra- and cross-dataset evaluation.


The remainder of this paper is organized as follows: firstly, in Section \ref{sec:related}, we conduct a comprehensive review of the rPPG approaches and the different methods used the estimate the HR in the existing literature. The proposed approach is presented in Section \ref{sec:method}, while experimental results are provided in Section \ref{sec:experiments}. Section \ref{sec:conclusions} summarizes our findings and conclusions.

\section{Related work}
\label{sec:related}
\subsection{rPPG measurement}


Since Takano et al. \cite{takano2007heart} and Verkruysse et al. \cite{verkruysse2008remote} demonstrated the feasibility of remote HR measurement from facial videos, researchers have proposed diverse methods for physiological data recovery. Some focus on regions of interest, utilizing techniques such as Blind Source Separation \cite{poh2010non, poh2010advancements, lewandowska2011measuring}, Normalized Least Mean Squares \cite{li2014remote}, or self-adaptive matrix completion \cite{tulyakov2016self}. Others leverage the skin optical reflection model by projecting RGB skin pixel channels into an optimized subspace, mitigating motion artifacts \cite{de2013robust, wang2015novel}.

In recent years, deep learning-based methods \cite{vspetlik2018visual, yu2019remote, perepelkina2020hearttrack, lee2020meta, lu2021dual, nowara2021benefit} have surpassed conventional techniques, achieving state-of-the-art performance in estimating vital signs from facial videos. Some combine knowledge from traditional methods with Convolutional Neural Networks (CNNs) to exploit sophisticated features \cite{niu2018synrhythm, niu2019rhythmnet, song2021pulsegan}. Recent works like \cite{lee2020meta, liu2021metaphys} explore unsupervised approaches using meta-learning, demonstrating improved generalization in out-of-distribution cases. On the other hand, some researchers focus on fully end-to-end approaches \cite{chen2018deepphys, Yu2019RemotePS, perepelkina2020hearttrack}. Unlike previous methods, end-to-end models use facial videos as input to predict the rPPG signal directly. Unsupervised techniques have gained traction in end-to-end approaches, with transformer-based models like Physformer \cite{yu2023physformer++} and RADIANT \cite{gupta2023radiant} leveraging long-range spatiotemporal features. However, these models currently lack optimization for computational efficiency, rendering them unsuitable for deployment on mobile devices. Additionally, they often depend on a prior fine-tuning stage due to limited generalization capabilities, especially in computer vision tasks \cite{dosovitskiy2021an}. While promising, they may not yet demonstrate a significant performance advantage over CNN-based models \cite{liu2023efficientphys}. Finally, works like \cite{liu2023efficientphys} and \cite{comas2022efficient} propose lightweight rPPG frameworks with competitive HR estimation results while controlling computational cost.
\subsection{HR estimation from rPPG}
\label{subsec:Heart rate estimation}

Once the rPPG signal is successfully extracted, a post-processing stage is employed to derive the corresponding HR value. Traditional literature distinguishes between temporal and frequency analysis. In the temporal domain, the peak detection method calculates the inter-beat intervals (IBI) between consecutive beats. Subsequently, HR is determined by averaging all IBIs over a time window and computing the inverse. Approaches \cite{poh2010advancements, balakrishnan2013detecting, mcduff2014improvements, wang2015novel} have heavily relied on these peak detector algorithms for HR estimation. Temporal HR analysis remains prevalent in state-of-the-art methods, especially in real-time applications like \cite{hansen2023real, gudi2020real}, or when intrinsic characteristics like HRV require analysis, as demonstrated in \cite{mcduff2014remote, yu2019remote}. However, these methods depend on accurate peak detection for robust instantaneous HR estimation, leading researchers to use customized peak detection functions for each presented database, making standardization challenging.

On the other hand, in frequency-based methods, the conventional approach involves converting the extracted rPPG signal to the frequency domain using the FFT, typically implementing Welch's method for density estimation. This technique assumes HR periodicity, where the highest power in the spectrum resides within the HR bandwidth. To mitigate the noise present in the rPPG signal, bandpass filtering in the HR range is commonly applied. This method, ranging from early rPPG handcrafted approaches \cite{poh2010non, lewandowska2011measuring, tulyakov2016self} to recent deep learning methods \cite{chen2018deepphys, perepelkina2020hearttrack, gupta2023radiant}, is widely adopted. However, a significant limitation of the FFT lies in assuming signal periodicity, restricting its frequency resolution. Consequently, the FFT may struggle to accurately discern closely spaced frequency components, especially in the narrowed HR band. Some attempts to address the weaknesses of the FFT were introduced by Irani et al. \cite{irani2014improved} and Yu et al. \cite{yu2013video}. The first one, suggested replacing the FFT with the Discrete Cosine Transform (DCT), demonstrating improved accuracy compared to FFT, while the second employed the Short-Fast-Fourier Transform (SFFT), proving more suitable for rapidly changing heart rate trends. However, these solutions only partially mitigate FFT issues. The DCT, being a derivative of the FFT, still suffers from the same FFT issues, while the SFFT involves a trade-off between temporal and frequency resolutions. Instead, in this paper, we propose a solution using the CZT, a generalization of the FFT that offers flexibility in specifying bandwidth without sacrificing good frequency and time resolution simultaneously.     

Despite many researchers still relying on traditional signal-processing techniques, the emergence of deep learning methods has led to various data-driven solutions. Among them, several researchers proposed the usage of deep HR regressors. Spetlik et al. \cite{vspetlik2018visual} pioneered this approach by combining a two-step convolutional neural network, with the second network being a deep HR estimator to extract the HR value. Niu et al. \cite{niu2019rhythmnet} introduced a deep regression model through a Gated Recurrent Unit (GRU), while works like \cite{wang2019vision, bousefsaf20193d, li2023non} incorporated fully connected layers into their end-to-end frameworks for direct HR extraction without prior rPPG signal recovery. Despite architectural differences, these methods share similar HR optimization using L1 loss or categorical cross-entropy loss. Even so, these optimization functions often force the model to learn periodic features within target frequency bands, which is challenging due to the present noise in the rPPG signal. Furthermore, the usage of categorical cross-entropy and the treatment of the HR band as a multiclass classification problem does not take into account the inter-class relationships between the different classes. To consider the ordinal structure of the HR values, we adopt a Squared Earth Mover’s Distance Loss \cite{hou2017squared} in the spectral density domain to optimize our deep CZT adaptative estimator. This loss aims to adapt the parameters of our deep CZT model to fit the characteristics of the HR sensor, which, jointly with the SMO loss, also guarantees a constrained adaptation, promoting generalization in cross-dataset evaluation.


\section{Methodology}
\label{sec:method}
In this section, we introduce and define the conventional CZT. Subsequently, we present our proposed deep CZT adaptative estimator and finally present our optimized objective function for remote HR estimation.

\begin{figure*}[h]
    \centering
\includegraphics[width=175mm,height=50mm]{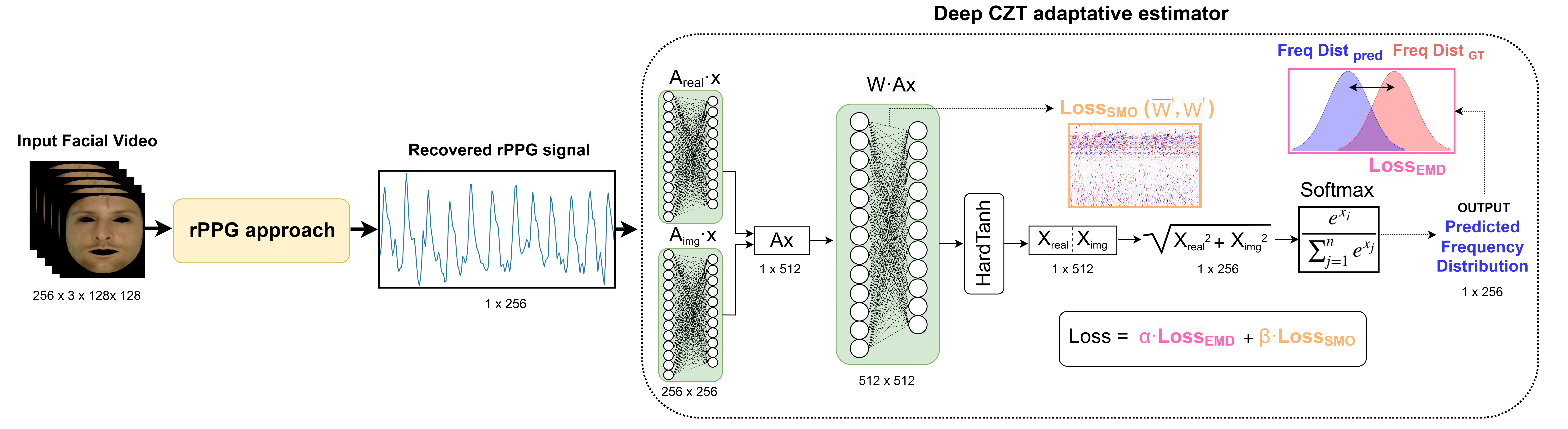}
    \caption{ Overall structure of our proposed deep CZT adaptative estimator model.} 
    \label{fig:CZTestimator}
\end{figure*}

\subsection{Chirp-Z Transform }

The CZT \cite{rabiner1969chirp} computes the z-transform of the finite duration signal $x[n]$ along a general spiral contour in the z-plane. Therefore, the CZT is defined  using the following formula:
\begin{equation}
\centering
CZT(x[n]) = \sum_{n=0}^{N-1} x[n] \cdot z_{k}^{-n}
\end{equation} 
Unlike the DFT, which evaluates the Z-transform of $x[n]$  on $N$ equally spaced points on the unit circle in the  z-plane, the CZT is not constrained to operate along the unit circle, evaluating the z-transform along spiral contours described as:
\begin{equation}
z_k = A \cdot W^{-k},  \quad  \text{\footnotesize{$k= 0,1, ..., M-1$}}
\end{equation}
\noindent where A is the complex starting point, W is a complex scalar describing the complex ratio between points on the contour, and M is the length of the transform. In addition, the CZT can also be expressed with the following matrix expression: 
\\\\
\begin{equation}
\renewcommand{\arraystretch}{1}
\setlength{\arraycolsep}{1.2pt}
\resizebox{0.915\linewidth}{!}{
    $\smash{\underbrace{\begin{bmatrix}
        X_0 \\ X_1 \\ X_2 \\ \vdots \\ X_n
    \end{bmatrix}}_{X}}
    = \smash{\underbrace{
    \begin{bmatrix}
        1 & 1 & 1 & \cdots & 1 \\
        1 & W^1 & W^2 & \cdots & W^{(n-1)} \\
        1 & W^2 & W^4  & \cdots & W^{2(n-1)} \\
        \vdots & \vdots & \vdots & \ddots & \vdots \\
        1 & W^{(m-1)} & W^{2(m-1)} & \cdots & W^{(m-1)(n-1)}
    \end{bmatrix}}_{W}}
    \smash{\underbrace{
    \begin{bmatrix}
        A^{-0} & 0 & 0 & \cdots & 0 \\
        0 & A^{-1} & 0 & \cdots & 0 \\
        0 & 0 & A^{-2}  & \cdots & 0 \\
        \vdots & \vdots & \vdots & \ddots & \vdots \\
        0 & 0 & 0 & \cdots & A^{-n}
    \end{bmatrix}}_{A}}
    \smash{\underbrace{\begin{bmatrix}
        x_0 \\ x_1 \\ x_2 \\ \vdots \\ x_n
    \end{bmatrix}}_{x}}$
}
\label{eq:matrixczt}
\end{equation}
\\

\noindent where A is a diagonal matrix $N$ by $N$ and W is an $M$ by $N$ Vandermonde matrix. 

During this work, we employ this matrix operation to compute the CZT, which is relevant for the design of our deep CZT adaptative estimator, described in subsection \ref{subsec:deepczt}. Although the CZT can be implemented in an efficient form such as FFT, by using the Bluestein algorithm \cite{bluestein1970linear}, we do not aim to find an efficient computation of CZT in this work. In this paper, we are interested in the CZT as an alternative form to estimate HR from rPPG signals by taking advantage of its spectral zoom property. Therefore, we propose to limit the CZT to the region of the unit circle because we are only interested in a narrow bandwidth of the rPPG frequency spectrum. So, we can define a zoom region that begins at $A$ and ends at $(M-1) \cdot W$. Following the literature and considering the majority of rPPG databases, we decided to limit the HR band 
within 0.66 and 3 Hz, equivalent to 40 to 180 beats-per-minute (BPM). Apart from defining the zoom spectrum region, the used bin density in the CZT is also configurable. After our preliminary experiments, we set the size of the CZT, $M$, equal to the size of the input size $N$, as shown in the example of Figure \ref{fig:FFTvsCZT}. This means, at the typical sampling rate of 30 frames per second, CZT has approximately 13 times higher frequency resolution than FFT in HR bandwidth, i.e. the same number of bins is used to cover $2.33$ Hz instead of 30 Hz.

\subsection{Deep CZT estimator}
\label{subsec:deepczt}

Our presented model, depicted in Figure \ref{fig:CZTestimator}, is designed to adapt the conventional CZT for estimating HR values from the rPPG signal. This adaptation involves making constrained modifications to the general structure of the CZT to learn the relationship between the input signal and the characteristics of the HR sensor. To achieve this, we represent the CZT framework as a neural network. However, addressing the complex nature of the transformation and ensuring proper optimization is challenging due to difficulties in the training dynamics between real and imaginary parts of the network, particularly at the backpropagation stage \cite{tan2022real}. We overcome this issue by relying on the structure of the CZT to avoid the direct use of the imaginary component. We duplicate the size of our $A$ and $W$ matrices, with the left half containing the real values and the right half containing the imaginary values, following Euler's formula:

 \begin{equation}
     \centering
     e^{ \pm i\theta } = \cos \theta \pm i\sin \theta
 \end{equation}

Applying this formulation in equation \ref{eq:matrixczt} we can express our deep CZT as the following matrix multiplication operation: 

\begin{equation}
\renewcommand{\arraystretch}{1.5}
\setlength{\arraycolsep}{2pt}
\resizebox{0.6\linewidth}{!}{
$\smash{\underbrace{\begin{bmatrix}
        X_{\RE} \\ X_{\IM}
    \end{bmatrix}}_{X}}
=
\underbrace{\left[\begin{array}{c;{2pt/2pt}c}
    W_{\RE} & -W_{\IM} \\ \hdashline[2pt/2pt]
    W_{\IM}  &   W_{\RE} \\
\end{array}\right]}_{W}
\smash{\underbrace{\begin{bmatrix}
    Ax_{\RE} \\ Ax_{\IM}
\end{bmatrix}}_{Ax}}$}
\label{eq:deepczt}
\end{equation}

Initially, the input rPPG signal $x$ is fed to two fully connected layers, which contain the weights of the $A$ matrix. One matrix for the real part (the $\cos()$ term), and another for the imaginary part (the $\sin()$ term). As $A$ is a diagonal matrix, and its diagonal denotes the starting frequency point, each fully connected layer contains a single parameter. Notably, this parameter is kept fixed, emphasizing optimization within a predetermined bandwidth between 0.66 and 3 Hz. Subsequently, adjustments are made to the $W$ matrix to accommodate complex number matrix multiplication between $W$ and $Ax$:
\begin{equation}
\begin{split}
    (W_{\RE} + i W_{\IM})(Ax_{\RE} + i Ax_{\IM}) =  (W_{\RE} \cdot Ax_{\RE}) - \\ (W_{\IM} \cdot Ax_{\IM}) + i\big((W_{\RE} \cdot Ax_{\IM}) + (W_{\IM} \cdot Ax_{\RE})\big)
\end{split}
\end{equation}
To optimize the $W$ matrix efficiently while preserving the CZT structure, we form the last fully connected layer as a square matrix initialized with the values of $W_\RE$
and $W_\IM$ from the classical CZT. To reduce the number of learnable parameters by half for $W$, we learn $W_\RE$ and $W_\IM$ once and then replicate them on the respective sides of the square $W$ matrix in Equation \ref{eq:deepczt}. As $W$ contains $\cos$ and $\sin$ values, and its weights fall within the range of -1 to 1, we constrain the learning of new weights to this range using a HardTanh function. Finally, we combine the real and imaginary components of the learned CZT by computing the modulus to obtain the learned frequential distribution. This distribution is then normalized using a Softmax function, and, similar to Welch’s method, the heart rate is computed by extracting the frequency with the maximal power response and multiplying it by 60.

\subsection{Loss function}
\label{subsec:loss_function}

To optimize the weights of the $W$ matrix in our deep CZT adaptive estimator, we employ a combined loss function. The aim of this combined loss function is two-fold: 1) Learn the closest frequency distribution from the rPPG signal to the distribution of the HR sensor used as ground truth. 2) Constrain the learning process to avoid capturing irrelevant noise features from specific databases, guiding the model to focus on meaningful adaptations of the classic CZT for remote HR estimation aligned with sensor-specific characteristics.

For the first objective, we expect the HR estimator to predict HR distributions where sub-bands closer to the HR ground truth distribution are assigned higher probabilities than those further away. Unlike Categorical Cross-Entropy or MAE loss, which do not consider the inter-class relationship in HR distributions, we propose an alternative approach using Earth Mover’s Distance (EMD). EMD is defined as the minimum cost to transport the mass of one distribution onto another one. To implement this, we adopt the Squared Earth Mover’s Distance-based Loss proposed in \cite{hou2017squared}, expressed as:
\begin{equation}
    \mathcal{L}_{EMD} = \frac{1}{N} \sum_{i=1}^{N} (CDF_{i}(p) - CDF_{i}(t))^2
\end{equation}
\noindent where $CDF()$ denotes the cumulative density function, $p$ and $t$ denote two compared distributions of same size $N$, which represents the batch dimensionality. In our framework,  $CDF(p)$ and $CDF(t)$ represent the frequency density functions of the predicted HR and the ground truth HR, as illustrated in Figure \ref{fig:CZTestimator}.

For the second objective, we seek a trade-off between the model customization of the HR sensor and the preservation of the classical CZT. Although the goal of our deep CZT adaptive estimator is to enhance the accuracy of HR estimation by learning the intrinsic features of the HR sensor, we aim to retain the structure of the classical CZT, which already demonstrates outstanding HR results, as will be shown in Section \ref{sec:experiments}. Therefore, to control the adaptation of the most relevant features, our $\mathcal{L}_\mathrm{{SMO}}$ acts as a regularization loss of the $ \mathcal{L}_{EMD}$ by constraining the learning of $W$. Referring to Subsection \ref{subsec:deepczt}, we focus exclusively on the $W$ learnable weights. Therefore, we construct a $\tilde{W}$ matrix comprising the $W_\RE$ and $W_\IM$ matrices without their replications needed for the full $W$ matrix: 

\newcommand{\zm}{%
    \left[\begin{array}{c;{2pt/2pt}c}
    W_{\RE} & W_{\IM} \\ 
    \end{array}\right]
}
\begin{equation}
\renewcommand{\arraystretch}{2}
\setlength{\arraycolsep}{3pt}
\resizebox{0.45\linewidth}{!}{
$ \tilde{W} = 
    \left.
  \,\smash[b]{\underbrace{\!\zm\!}_{\textstyle\text{$N$}}}\,
  \right\}\text{$M$}
  \vphantom{\underbrace{\zm}_{\text{$n$}}}
$
}
\end{equation}
Here, $N$ represents the number of columns, which is twice the input signal's size after separating the real and imaginary components, and $M$ denotes the number of rows, where each row corresponds to a frequency sub-band between 0.66 and 3 Hz. To control the flexibility in learning the optimized $W$ through the $ \mathcal{L}_{EMD}$, we perform HR frequency sub-band optimization. This involves comparing each row value between $\tilde{W}'$ and $\tilde{W}$, subsequently averaging all the sub-band errors, expressing $\mathcal{L}_\mathrm{{SMO}}$ as:


\begin{equation}
    \mathcal{L}_\mathrm{{SMO}} = \frac{1}{L} \sum_{i=1}^{M}\sum_{j=1}^{N} |\tilde{W}'(i, j) - \tilde{W}(i,j)|
\end{equation}


\noindent where $\tilde{W}'$ represents the learnable version of $\tilde{W}$ from our CZT estimator and $L$ denotes the total number of frequency bins. In summary, the overall loss function $\mathcal{L}_\mathrm{{HR}}$ can be formulated as:
\begin{equation}
\label{combined_eq}
    \mathcal{L}_\mathrm{{HR}} = \alpha \cdot \mathcal{L}_\mathrm{{EMD}} + \beta \cdot \mathcal{L}_\mathrm{{SMO}},
\end{equation}
 where $\alpha$ and $\beta$ are balancing parameters. In our experiments, we set $\alpha = 100$ and $\beta = 0.01$ empirically based on our preliminary experiments. These settings yield a balanced contribution of both losses, taking into account that the magnitude of $\mathcal{L}_\mathrm{{SMO}}$ is much larger than that of $\mathcal{L}_\mathrm{{EMD}}$.

\section{Experiments}
\label{sec:experiments}
In this section, we introduce three benchmark datasets and outline the implementation of our methodology. Our analysis begins by comparing HR estimations derived from PPG ground truth and sensor HR data. Specifically, in the initial experiment, we compare the classical CZT with baseline methods for HR estimation, concurrently assessing the influence of signal length. Subsequently, we evaluate remote HR estimation performance, referencing the ground truth data. Our exploration then extends to examining the impact of the loss function in our deep CZT model. We conduct a comprehensive evaluation through both intra-database and cross-database assessments, showcasing its robustness in the context of remote HR estimation. Finally, we test our deep CZT model with various rPPG models, highlighting its versatile capability with different rPPG approaches.

\subsection{Datasets}
\label{subsec:datasets}
We assessed our approach on the following rPPG datasets, since each of them contains PPG and HR ground truth data.

The UCLA-rPPG dataset \cite{wang2022synthetic} comprises 489 videos from 98 subjects with diverse characteristics, including skin tones, ages, genders, and ethnicities. 
Each subject underwent five trials, with each trial lasting approximately 1 minute. The recordings were captured at a resolution of $640\times480$ pixels and 30 frames per second (FPS), in an uncompressed format. Synchronous gold-standard PPG and HR measurements were collected alongside the facial videos. Due to the lack of predefined folds in this dataset, we split the data into training (80\%), validation (10\%), and testing (10\%) sets. 

The UBFC-rPPG \cite{bobbia2019unsupervised} includes 42 RGB videos from 42 subjects. The subjects were asked to play a time-sensitive mathematical game, emulating a standard human-computer interaction scenario, to obtain varied HR during the experiment. The recorded facial videos were acquired indoors with varying sunlight and indoor illumination at 30 FPS with a webcam (Logitech C920 HD Pro) at a resolution of 640x480 in uncompressed 8-bit RGB format. The bio-signals ground truth was acquired using a CMS50E transmissive pulse oximeter to record the PPG signal and heart rate. In our experiments, we used UBFC-rPPG in a cross-dataset evaluation, where all 42 videos were used only for testing.

The Pulse Rate Detection Dataset (PURE) contains 60 videos from 10 subjects (8 male, 2 female) performing six different head motion tasks: steady, talking, slow translation, fast translation, small rotation, and medium rotation. The facial videos were recorded using an ECO274CVGE camera with a resolution of 640 x 480 pixels and 30 FPS. Each video is about 1 minute long and stored in uncompressed PNG format. The gold-standard measures of BVP and SpO2 were collected using a finger pulse oximeter. Similar to the UBFC-rPPG dataset, we considered the PURE dataset only for cross-dataset evaluation.

\begin{figure*}[t]
    \centering
\includegraphics[width=130mm,height=40mm]
{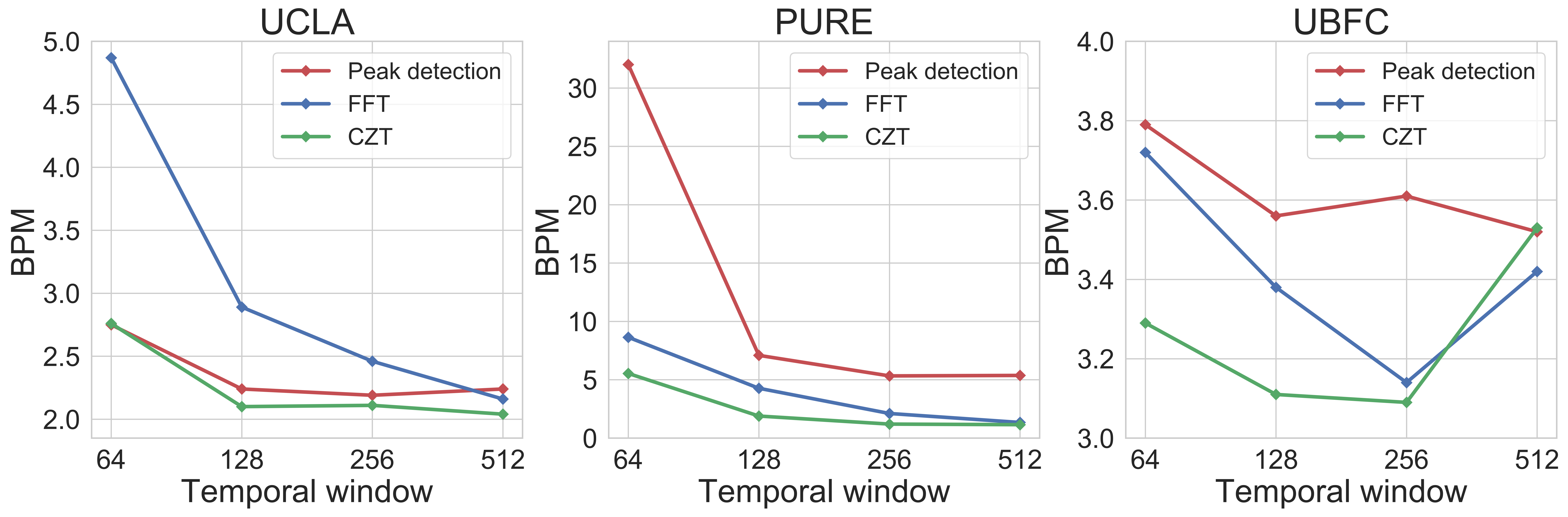}
    \caption{MAE comparison between handcrafted HR estimation methods using different temporal windows with ground truth data.}
    \label{fig:mae_seq}
\end{figure*}

\subsection{Implementation details}
\subsubsection{Preprocessing and training procedure}
In all our experiments, we adopt the preprocessing stage from \cite{comas2022efficient} for each dataset. However, a minor modification is made to facial region estimation compared to \cite{comas2022efficient}. Instead of using the MTCNN \cite{zhang2016joint} algorithm, we choose for facial preprocessing segmentation, utilizing the Mediapipe Face Mesh\footnote{\url{https://github.com/google/mediapipe/blob/master/docs/solutions/face_mesh.md}} model \cite{kartynnik2019real}. This adjustment simplifies rPPG estimation by focusing on facial skin without introducing additional complexity in terms of parameters or optimization tasks. After masking the facial video, each frame is resized to $128\times 128$ pixels.

Our deep CZT estimator module undergoes a two-stage training process. Initially, we pre-train all rPPG approaches used in subsections \ref{subsec:ablation} and \ref{subsec:sota}, adhering to the specifications of each approach following \cite{liu2022deep} and \cite{comas2022efficient}.  Subsequently, we train our CZT estimator by initializing all weights similar to the traditional CZT, while keeping the weights of the rPPG method frozen. The implementation uses PyTorch 1.8.0 \cite{paszke2019pytorch} and is trained on a single NVIDIA GTX1080Ti. Sequences of 256 frames without overlap are used during training. We employ the AdamW optimizer with a learning rate of 0.0001 and a ReduceLROnPlateau learning rate scheduler with a patience of 5 and a factor of 0.9. Regarding the experiment reproducibility, the details of our experimental setup, as presented in Section \ref{sec:experiments}, will be made publicly available in a GitHub repository.

\subsubsection{Metrics and evaluation}      

To evaluate the HR estimation performance of the proposed model, we adopt the same metrics used in the literature, such as the mean absolute HR error (MAE), the root mean squared HR error (RMSE), the mean absolute percentage error (MAPE) and Pearson’s correlation coefficients R \cite{li2014remote, liu2021metaphys, liu2023efficientphys}. Ablation experiments are performed using 256-frame sequences (8.5 seconds approx) with no overlap to compute HR estimation for all reported metrics, which is more challenging and informative than HR estimation based on the whole video sequence at once. Importantly, our goal is to approximate the HR sensor's value, and thus, all reported results are compared against the HR sensor ground truth data of each dataset without deriving HR from the PPG signal, as is commonly done in other works more focused on assessing the quality of the rPPG signal than on the estimated HR.


\subsection{Preliminary experiment: Why CZT for remote HR estimation?}
\label{subsec:preliminary}
In our preliminary experiment, we aim to highlight the advantages of employing the CZT for remote HR estimation. This initial study has two main objectives: 1) to quantify the improvements in accuracy with respect to standard practice, and 2) to show the possibility of shorter-window estimates.

Firstly, we estimate and compare HR values using directly the PPG signal provided by the evaluated datasets (i.e. simulating a perfect rPPG estimation), benchmarking against two widely used techniques for HR estimation: peak detection and FFT. Secondly, we evaluate the performance of each technique across varying temporal windows for each dataset, investigating the impact of temporal window size on HR estimation.

In Figure \ref{fig:mae_seq}, the MAE between the HR extracted from the PPG ground truth signal and HR ground truth data is depicted for UCLA-rPPG, PURE, and UBFC-rPPG datasets. The behaviour of each HR estimation method is explored across four temporal windows (2.5 to 17 seconds, or 64 to 512 samples), incremented in powers of two, for each dataset. A consistent trend emerges across datasets, showing HR error decreasing with larger window sizes, as shorter temporal windows pose greater challenges. Comparing the three methods, the CZT (in green) achieves the lowest error in the three datasets across most window sizes, maintaining a similar error across different temporal windows. We observe two important deviations from these general observations:

\begin{enumerate}
    \item In the UBFC-rPPG dataset for a window size of 512 samples, the HR error for both frequential methods increases instead of reducing with respect to the error using shorter windows. This is due to the high variability of HR over time for this dataset, as subjects were playing a mathematical game while recorded. In such case, estimating HR based on a unique maximum of the spectrum, as done in FFT and CZT, inevitably leads to inaccuracies. 
    \item In the PURE dataset for a window size of 64 samples, we observe that the temporal method (in red) produces comparatively large errors. This happens because the majority of recordings in the PURE dataset are for subjects with HR between 40 and 60 BPM; therefore there are very few peaks in such a short temporal window and the resulting estimates are poor.
\end{enumerate}


A notable observation from this preliminary experiment is that none of the employed methods yields an absolute error of zero in any dataset. While in the ideal scenario, we anticipate a perfect match between the PPG signal and the HR values acquired from the HR sensor, this initial experiment reveals an error gap between both ground truth signals. Even with a highly precise rPPG approach, an error persists in the recovered signal and the HR values \cite{gudi2020real}. Determining the cause of this difference is challenging, given that most oximeter devices do not provide information on how HR is estimated from the PPG signal, and this process may vary across devices. This underscores the importance of establishing a mapping between the recovered PPG signal and the HR value, which is the objective of our deep CZT estimator, studied in the next section.

\subsection{Ablation studies}
\label{subsec:ablation}
In this subsection, we will evaluate our proposed deep CZT estimator module. For our ablation studies we choose as rPPG approach the TDM model \cite{comas2022efficient}, because it consists of a lightweight method that obtains competitive rPPG results controlling the amount of parameters.

\begin{table}
\caption{Comparison of different loss functions (beats per minute).}
\renewcommand{\arraystretch}{1.15}
\centering
\adjustbox{width=0.49\textwidth}{
  \begin{tabular}{c|c c c c}
    \hline
    \multirow{2}{0.3cm}{Loss} & \multicolumn{4}{c}{Intra-database evaluation }  \\ [0.2ex]
    \cline{2-5}
    & MAE$\downarrow$&RMSE$\downarrow$&MAPE$\downarrow$&R$\uparrow$\\
    \hline
    $\mathcal{L}_\mathrm{{CE}}$ & 2.22$\pm$0.28 & 2.98$\pm$2.18 & 2.91$\pm$0.35 & \textbf{0.97$\pm$0.03}\\[0.2ex]
    $\mathcal{L}_\mathrm{{EMD}}$  &  1.88$\pm$0.32 & 2.94$\pm$2.90 & 2.45$\pm$0.38  & 0.96$\pm$0.04\\ [0.2ex]
    $\mathcal{L}_\mathrm{{EMD}}+\mathcal{L}_\mathrm{{SMO}}$ & \textbf{1.74$\pm$0.26} & \textbf{2.53$\pm$1.67} & \textbf{2.26$\pm$0.32} & \textbf{0.97$\pm$0.03} \\ 
    \hline
    \end{tabular}
    }
    \label{table:loss_exp}
    \end{table}

\subsubsection{Loss function impact}

In Table \ref{table:loss_exp}, we perform a loss function experiment for our deep CZT estimator, testing different losses in intra-database evaluation. As detailed in subsection \ref{subsec:deepczt}, our proposed model outputs the estimated frequency distribution. Therefore, we explore three distinct losses: cross-entropy, squared Earth Mover’s Distance-based loss, and our proposed loss, which combines the Squared Earth Mover’s Distance-based Loss with SMO regularization loss. 

Comparing the cross entropy and the squared EMD loss, we can appreciate a significant difference, especially in MAE and MAPE metrics where the error is considerably lower. As explained in section \ref{subsec:Heart rate estimation}, this discrepancy can be attributed to the cross-entropy loss not considering inter-class relationships among different HR classes, whereas the EMD loss incorporates the ordinal behavior of the HR frequency distribution.
Finally, we find the performance of our combined loss, $\mathcal{L}_\mathrm{{EMD}}+\mathcal{L}_\mathrm{{SMO}}$, which denotes the best HR results for all the metrics surpassing the previous EMD and cross-entropy losses. This superior performance can be attributed to the regularization 
imposed over the parameters of the matrix W, compelling our estimator to modify only relevant information. This preserves the traditional CZT structure, preventing the learning of inappropriate modifications or overfitting.

\begin{figure}[t]
    \centering
\includegraphics[width=80mm,height=65mm]{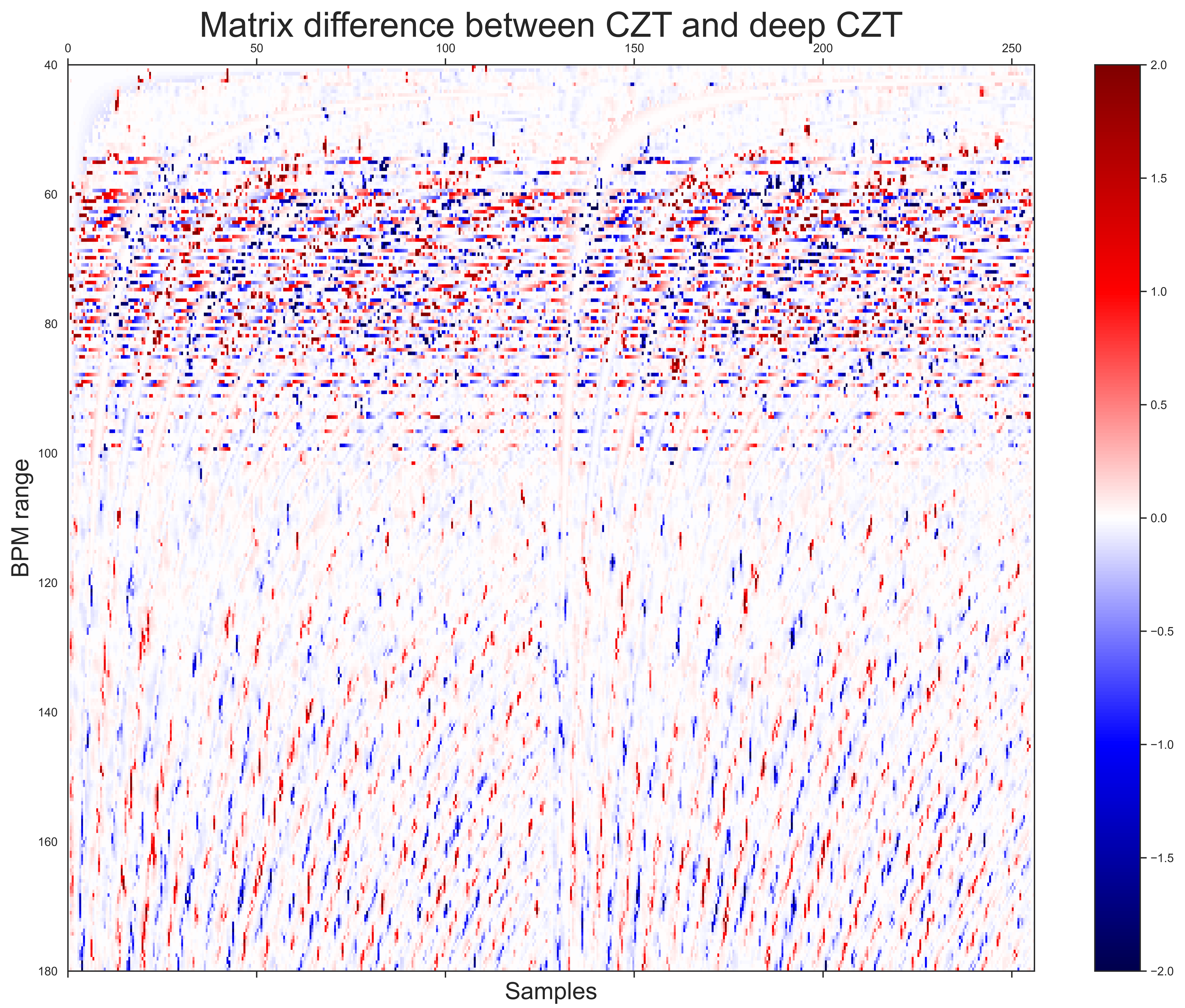}
    \caption{Sparse Matrix optimization result. Difference between $\tilde{W}$ matrix of standard CZT and $\tilde{W}'$ matrix of trained deep CZT estimator.}
    \label{fig:matrix_diff}
\end{figure}

In Figure \ref{fig:matrix_diff}, the SMO results are illustrated, depicting the difference map between the initial $\tilde{W}$ matrix of the standard CZT and our learned $\tilde{W}'$ matrix after training our CZT estimator. The majority of changes, fluctuating between $-2$ and $+2$, mainly occur in the frequency bands between 50 and 100 BPM (on the y-axis), aligning with the central distribution of the UCLA-rPPG dataset around 60-70 BPM, which is the training dataset. Despite observing more changes across the entire matrix, the color is more attenuated, indicating that the difference between the original and learned weights is nearly negligible.

\subsubsection{Cross-database evaluation}
\label{subsubsec:cross-eval}
In Table \ref{table:cross_eval}, we evaluate our deep CZT estimator's performance with the proposed combined loss in cross-dataset scenarios, specifically in the PURE and UBFC-rPPG datasets. The comparison includes the FFT and the standard CZT for a comprehensive analysis. 

In the PURE dataset, we can see a notable difference between the traditional FFT and the CZT, where the MAE and RMSE errors are 1 BPM and 2.5 BPM lower, respectively. Regarding deep CZT, we obtain competitive results even in cross-dataset evaluation, improving the HR error from the traditional methods, highlighting an RMSE of 6.11 BPM or a MAPE of 5.23 \%. On the other hand, in the UBFC-rPPG dataset, CZT and FFT exhibit more similar errors. CZT secures better results in MAE and MAPE, while FFT shows slightly better RMSE and R. In contrast, the deep CZT estimator significantly enhances performance across all evaluated metrics, with an RMSE of 6.58 and a MAPE of 4.36\%, indicating a more robust HR estimation.

\begin{table}[b]
  \caption{Cross-database heart rate evaluation on PURE and UBFC-rPPG (beats per minute).}
  \renewcommand{\arraystretch}{1.35}
  \centering
  \adjustbox{width=0.49\textwidth}{
  \begin{tabular}{c|c|cccc}
    \hline
    \multirow{2}{*}{Dataset} & \multirow{2}{*}{HR Method} & \multicolumn{4}{c}{Cross-database evaluation}\\ 
    \cline{3-6}
    && MAE$\downarrow$ & RMSE$\downarrow$ & MAPE$\downarrow$ & R$\uparrow$ \\ 
    \hline
        \multirow{3}{*}{{\parbox{1cm}{\centering PURE}}} & FFT & 4.34$\pm$1.23 & 10.37$\pm$69.21 & 7.32$\pm$2.36 & 0.90$\pm$0.06 \\
        & CZT & 3.41$\pm$0.93 & 7.94$\pm$34.08 & 5.84$\pm$1.79 & 0.94$\pm$0.05 \\
        & Deep CZT & \textbf{3.37$\pm$0.66} & \textbf{6.11$\pm$11.54} & \textbf{5.23$\pm$1.11} & \textbf{0.97$\pm$0.03}  \\
    \hline
        \multirow{3}{*}{{\parbox{1cm}{\centering UBFC}}} 
        & FFT & 4.83$\pm$0.94 & 7.76$\pm$20.97& 5.40$\pm$1.12& 0.89$\pm$0.07 \\  
        & CZT &4.66$\pm$0.97 & 7.84$\pm$21.21& 5.07$\pm$1.09& 0.88$\pm$0.07 \\
        & Deep CZT & \textbf{4.19$\pm$0.78}& \textbf{6.58$\pm$12.99}& \textbf{4.36$\pm$0.82}& \textbf{0.92$\pm$0.06}\\
    \hline
  \end{tabular}
  }
  \label{table:cross_eval}
\end{table}

\begin{table*}[t]
  \caption{Performance evaluation of deep CZT estimator in SOTA approaches  (beats per minute).}
  \renewcommand{\arraystretch}{1}
  \centering
  \adjustbox{width=0.68\textwidth}{
  \begin{tabular}{c|c|cccc}
    \hline
    \multirow{3}{2cm}{\centering rPPG Approach} & \multirow{3}{*}{HR Method} &\multicolumn{4}{c}{Intra-database evaluation} \\ 
    \cline{3-6} && \multicolumn{4}{|c}{UCLA-rPPG}\\ 
    \cline{3-6}
    &&MAE$\downarrow$&RMSE$\downarrow$&MAPE$\downarrow$&R$\uparrow$\\ 
    \hline
    \multirow{3}{*}{{\parbox{2cm}{\centering LGI \cite{pilz2018local}}}} & 
    FFT& 3.28$\pm$0.47& 4.64$\pm$6.21& 4.24$\pm$0.54& 0.93$\pm$0.05\\  
    & CZT & 2.88$\pm$0.38& 3.92$\pm$3.67&3.71$\pm$0.44&\textbf{0.96$\pm$0.04}\\
    & Deep CZT & \textbf{2.40$\pm$0.32}&\textbf{3.29$\pm$2.51}&\textbf{3.12$\pm$0.38}& \textbf{0.96$\pm$0.04} \\
    \hline
    \multirow{3}{*}{{\parbox{2cm}{\centering POS \cite{wang2016algorithmic}}}} & 
    FFT& 3.17$\pm$0.50& 4.71$\pm$8.18& 4.40$\pm$0.76& 0.87$\pm$0.07\\  
    & CZT & 2.62$\pm$0.34& 3.56$\pm$2.67& 3.60$\pm$0.50&0.93$\pm$0.05\\
    & Deep CZT & \textbf{2.16$\pm$0.30}&\textbf{3.00$\pm$2.10}&\textbf{2.87$\pm$0.38}& \textbf{0.97$\pm$0.04} \\
    \hline
    \multirow{3}{*}{{\parbox{2cm}{\centering Physnet \cite{Yu2019RemotePS}}}} & 
    FFT&2.27$\pm$0.28&2.99$\pm$1.76&2.96$\pm$0.34&\textbf{0.98$\pm$0.03}\\
    &CZT&2.26$\pm$0.30&3.10$\pm$2.21&2.91$\pm$0.37&\textbf{0.98$\pm$0.03}\\
    &Deep CZT & \textbf{1.78$\pm$0.24}&\textbf{2.47$\pm$1.59}&\textbf{2.32$\pm$0.30}&\textbf{0.98$\pm$0.03}\\
    \hline
    \multirow{3}{*}{{\parbox{2cm}{\centering Efficient-Phys \cite{liu2023efficientphys}}}} & 
    FFT & 2.73$\pm$0.45&4.17$\pm$6.21&3.73$\pm$0.67&0.91$\pm$0.06\\ 
    & CZT & 2.13$\pm$0.34&3.17$\pm$3.23&2.76$\pm$0.41& 0.95$\pm$0.04\\
    & Deep CZT & \textbf{1.68$\pm$0.26}&\textbf{2.45$\pm$1.61}&\textbf{2.22$\pm$0.32}&\textbf{0.97$\pm$0.03}\\
    \hline
    \multirow{3}{*}{{\parbox{2cm}{\centering TDM \cite{comas2022efficient}}}} & 
    FFT & 2.50$\pm$0.34&3.43$\pm$3.02&3.25$\pm$0.41&0.95$\pm$0.04 \\  
    & CZT &2.27$\pm$0.30&3.08$\pm$2.23&2.93$\pm$0.36&\textbf{0.97$\pm$0.03} \\
    & Deep CZT & \textbf{1.74$\pm$0.26}&\textbf{2.53$\pm$1.67}&\textbf{2.26$\pm$0.32}&\textbf{0.97$\pm$0.03}\\
    \hline
  \end{tabular}
  }
  \label{table:sota_adaptation}
\end{table*}

\subsection{Adapting deep CZT estimator to SOTA rPPG approaches}
\label{subsec:sota}

Despite SMO regularization aiding intra-dataset optimization, the primary goal is to preserve the CZT structure. As indicated in subsection \ref{subsec:preliminary}, the standard CZT outperforms commonly used techniques in the literature. Therefore, while SMO regularization adjusts weights for intra-dataset characteristics, it also preserves generalization ability across datasets.

In this section, we assess the performance of our HR estimator module using five selected rPPG methods: POS \cite{wang2016algorithmic}, LGI \cite{pilz2018local}, PhysNet \cite{Yu2019RemotePS}, EfficientPhys \cite{liu2023efficientphys} (implemented from the rPPG-toolbox \cite{liu2022deep}), and the TDM model \cite{comas2022efficient}, previously employed in subsection \ref{subsec:ablation}. Similarly to the previous section, we compare our deep CZT estimator against FFT and CZT, used as benchmarks for each method's results, as summarized in Table \ref{table:sota_adaptation}.

Results consistently align with the findings presented in our preliminary experiment (Section IV.C), showing that CZT tends to accurately extract the HR value with lower error compared to FFT. Among the five methods, CZT exhibits lower HR error in all metrics except in the Physnet model, where the performance is nearly the same with very small differences. On the other hand, our adaptable deep CZT estimator shows a significant reduction in HR error not only compared to FFT but also compared to the standard CZT. 

In terms of rPPG estimation, the use of our deep CZT in traditional methods such as LGI or POS, improves their HR performance by reducing the MAE and the RMSE between 0.5 and 1.71 BPM. A similar trend is observed for data-driven methods: using our deep CZT estimator outperform previous results, reducing the MAE and the RMSE below 2 and 3 BPMs, respectively. These promising results indicate that our deep CZT estimator successfully narrows the gap between the HR extracted from the rPPG signal and the HR value from the dataset sensor. Moreover, it proves to be independent of the adopted rPPG method, whether traditional or data-driven, highlighting its potential to be incorporated into any rPPG model designed for HR estimation. 

\section{Conclusions}
\label{sec:conclusions}


This paper proposes the use of the CZT for remote heart rate measurement. Its inherent flexibility in adjusting frequency resolution overcomes the limitations of the FFT, enabling more robust and precise HR estimation across various temporal window sizes. Additionally, we introduce a novel data-driven CZT estimator tailored to adapt the classical CZT to the unique characteristics of each HR sensor and the recovered PPG signal.

To guide the adaptation of our model, we propose a frequency distribution loss regularized with sparse matrix optimization, showcasing outstanding results in both intra-database and cross-dataset evaluations. Furthermore, we validate the capability of incorporating our deep CZT estimator into several existing rPPG methods, highlighting its adaptability and improved performance compared with current frequency handcrafted methods.

In conclusion, we have presented an alternative to the FFT for remote heart rate estimation, leveraging the CZT. Our results indicate promising performance, even when employing temporal windows as short as approximately 2 seconds. This implies a comparatively lower delay in the estimate of heart rates and suggests the possibility of exploring applications that may require near-instantaneous HR estimation.
\section*{Acknowledgments}
This work is partly supported by the eSCANFace project (PID2020-114083GB-I00) funded by the Spanish Ministry of Science and Innovation.

{\small
\bibliographystyle{abbrv}
\bibliography{References}
}

\end{document}